\documentclass[final]{neus2025}
\usepackage{graphicx}

\usepackage{tikz}
\def\checkmark{\tikz\fill[scale=0.4](0,.35) -- (.25,0) -- (1,.7) -- (.25,.15) -- cycle;}

\usepackage{algorithm}
\usepackage{algpseudocode}

\title[ChatHTN]{ChatHTN: Interleaving Approximate (LLM) \\
 and Symbolic HTN Planning}
\usepackage{times}

\author{%
 \Name{H\'ector Mu\~noz-Avila} \Email{hmunozavila@american.edu}\\
 \addr American University
 \AND
 \Name{David W. Aha} \Email{david.w.aha.civ@us.navy.mil}\\
 \addr Navy Center for Applied Research in AI; Naval Research Laboratory; Washington, DC%
\AND
 \Name{Paola Rizzo} \Email{p.rizzo@interagens.com}\\
 \addr Interagens s.r.l.%
}

\begin{document}

\maketitle

\begin{abstract}

We introduce ChatHTN, a Hierarchical Task Network (HTN) planner that combines symbolic HTN planning techniques with queries to ChatGPT to approximate solutions in the form of task decompositions. The resulting hierarchies interleave task decompositions generated by symbolic HTN planning with those generated by ChatGPT. Despite the approximate nature of the results generates by ChatGPT, ChatHTN is provably sound; any plan it generates correctly achieves the input tasks. We demonstrate this property with an open-source implementation of our system.

\end{abstract}

\bigskip


\section{Introduction}

Hierarchical Task Network (HTN) planning is a paradigm in which high-level tasks are decomposed into simpler ones. HTN planning has been the subject of frequent studies because many applications are amenable to hierarchical modeling, including military planning \cite{Donaldson2014}, strategic decision making (e.g., in games \cite{smith1998success,Verweij2007}), and controlling multiple agents \cite{cardoso2017multi}, including teams of UAVs \cite{musliner2010priority}. Another reason for this recurrent interest are studies reporting that hierarchical modeling is a natural way for automated systems to learn skills of increasing complexity (i.e., by starting with simpler skills and then combining them to learn more complex skills \cite{langley2006learning}). Indeed, hierarchical planning is a key component in many cognitive architectures \cite{laird2019soar,langley2006unified} as they aim to create agents that are endowed with advanced cognitive capabilities.

LLMs cannot generate plans that are guaranteed to be \textit{sound}. That is, they cannot guarantee that the solutions they generate correctly solve given planning problems \cite{valmeekam2024llms}. Furthermore, they cannot verify that generated plans are indeed solutions for given problems \cite{valmeekam2024planningstrawberryfieldsevaluating}. Noticeably, they \textit{can} generate plans for problems that have no solution (e.g., because a key resource is not given in the problem description). Nevertheless, they excel at generating \textit{plan approximations} (i.e., plausible plans for solving problems for task domains an LLM has observed in its training data). Since they are trained with vast volumes of information, exploiting these plan approximations is an enticing possibility.

In this paper, we introduce ChatHTN, which combines two ideas: (1) HTN planning with soundness guarantees and (2) approximate hierarchical planning. In a nutshell, ChatHTN is an HTN planner: it maintains a task list, indicating the pending tasks it needs to accomplish, and uses methods and operators to achieve these tasks. The task list is symbolic; each task is represented as a ground predicate (i.e., a predicate whose arguments are all constants). If ChatHTN attempts to decompose a task for which it has no task decomposition in its knowledge base, then it asks ChatGPT for a plausible decomposition of the task and then reverts back to HTN planning.

We have two objectives in introducing ChatHTN. First, we aim to alleviate one of the most stringent requirements of HTN planning: that a complete knowledge base is given. That is, knowledge (in the form of methods) must be provided to decompose every task in every situation  where a decomposition is possible. This is an instance of the well documented knowledge engineering bottleneck that has been a recurrent, long-term thorn for AI systems. Second, we address the inability of LLMs to generate sound plans. This is also a crucial obstacle for the reliable use of LLMs, particularly in the context of mission critical, fail-safe environments such as self-driving vehicles, automated flight assistants, and power grid control. 

\section{Motivating Example}

Consider the logistics transportation domain. In this domain, the main task is to relocate a package. There are trucks that can be used to move packages between locations provided that their locations are in the same city. There are also airplanes that can be flown between any two airports.

\begin{figure}[htb]
    \centering
    \includegraphics[width=14cm]{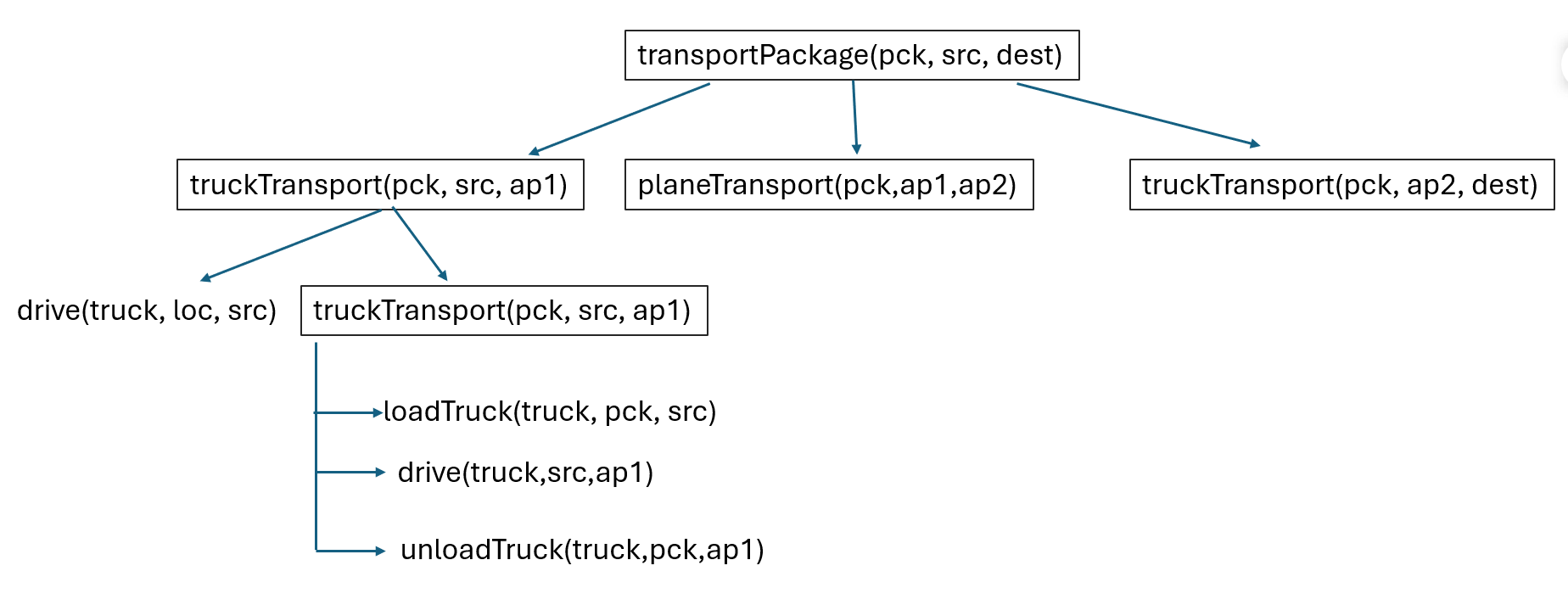}
   \caption{Initial hierarchical decomposition for a task from the logistics transportation domain. Boxed predicates are compound tasks and unboxed tasks are primitive tasks. ChatHTN stops on the planeTransport task because it lacks the knowledge to decompose it.}
    \label{fig:exampleStart}
\end{figure}

Consider a task \textit{transportPackage(pck, src, dest)}  to relocate package \textit{pck} from the source location to the destination location. ChatHTN proceeds with standard HTN planning and generates the hierarchy shown in Figure \ref{fig:exampleStart}. The \textit{transportPackage} task is decomposed into 3 compound tasks: truckTransport, planeTransport and truckTransport (with parameters as shown in the figure). It decomposes the first truckTransport task, all the way to primitive tasks. However, it stops at the planeTransport task because it has no means to decompose this task given the current state and knowledge base.

\begin{figure}[htb]
    \centering
    \includegraphics[width=14cm]{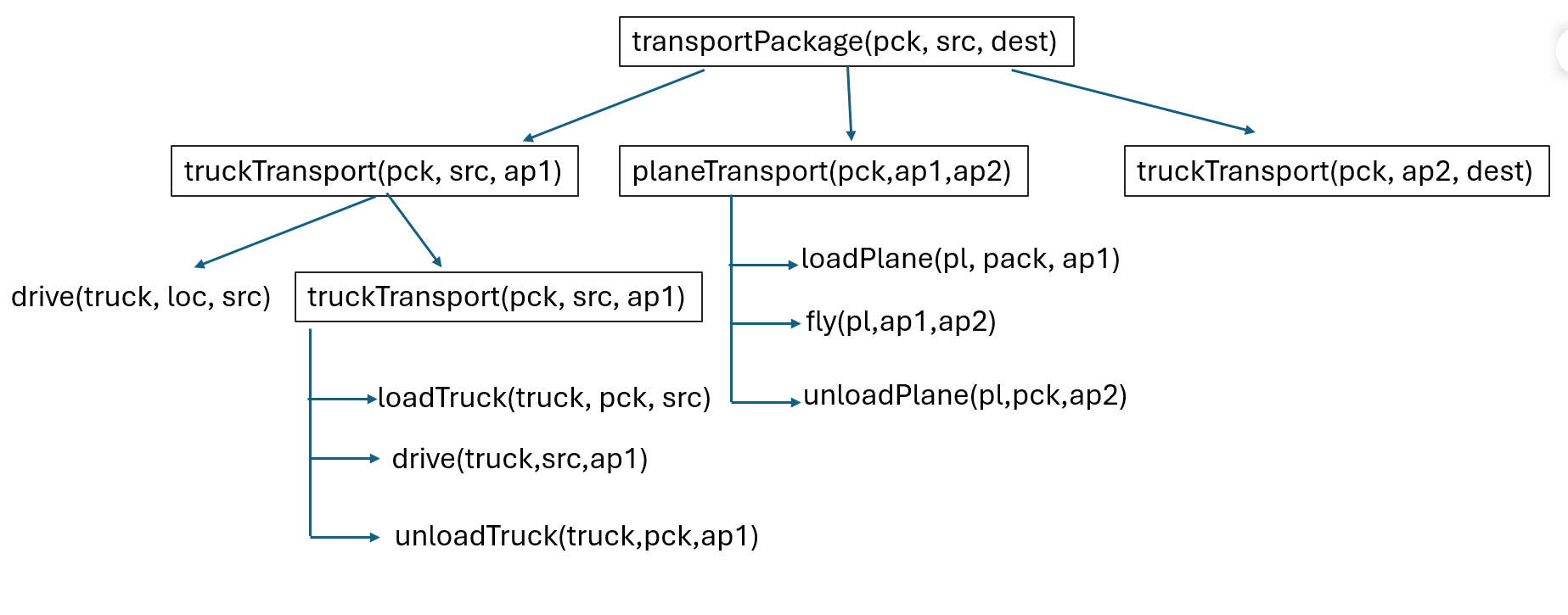}
   \caption{ChatHTN  prompts ChatGPT for a decomposition of  the planeTransport task.}
    \label{fig:exampleMiddle}
\end{figure}

\begin{figure}[htb]
    \centering
    \includegraphics[width=14cm]{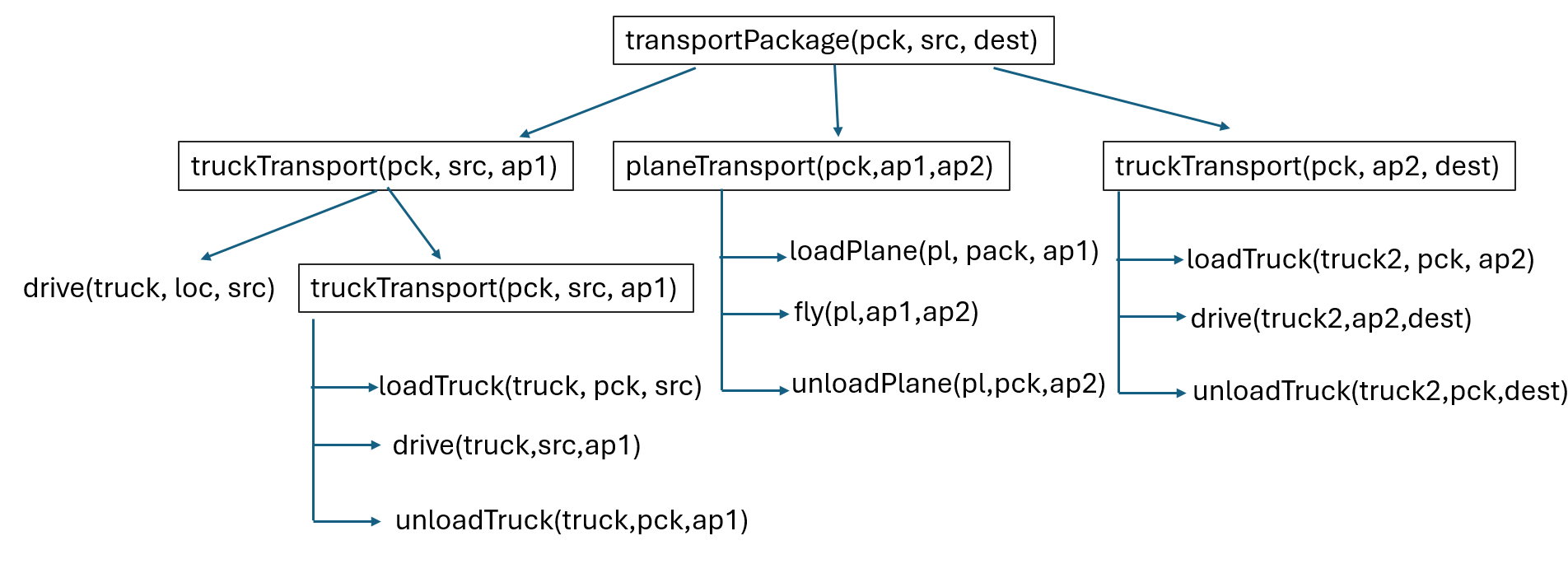}
   \caption{ChatHTN continues hierarchical planning for the remaining task to generate a complete plan.}
    \label{fig:exampleFinal}
\end{figure}

In this case, a standard HTN would return $\emptyset$ (signifying that no solutions were found). ChatHTN instead uses ChatGPT to generate a possible decomposition for this task. In doing so, ChatHTN provides ChatGPT with additional information for context such as the task's semantics, the state, and part of the knowledge base. ChatGPT is prompted to generate a decomposition of this task into a sequence of primitive tasks.  ChatGPT returns the decomposition displayed in Figure \ref{fig:exampleMiddle} for the \textit{planeTransport} task. Then ChatHTN continues the standard HTN planning decomposition process. A complete HTN is shown in Figure \ref{fig:exampleFinal}.

\section{Terminology}

The first three rows of Table 1 describe \textit{compound} tasks; they are decomposed into other tasks but will not change the state of the world. In the standard definition of HTN planning, compound tasks have no preconditions and effects \cite{erol1994htn}. However, some work on learning HTN planning knowledge does assign semantics as this is needed to guide the learning system \cite{langley2009learning}. Here we define tasks as \textit{(preconditions,effects)} pairs, as in \cite{hogg2008htn}, which allows ChatHTN to provide ChatGPT with the semantics of the tasks to be decomposed. For instance, the \textit{truckTransport(p, s, d)} task (third row) requires that package \textit{p} to be at location \textit{s} and the destination location \textit{d} to be in the same city as location \textit{s}. The expected effect of the task is for \textit{p}  to be at \textit{d}.\footnote{Table 1 displays only the main preconditions for the sake of space. Missing conditions include checking that the arguments are of the correct type, such as \textit{isPackage(p)}, which checks that $p$ is a valid package. We do the same for the effects.}

Compound tasks express \textit{what} needs to be accomplished but not \textit{how}. To decompose a task, \textit{methods} are used. They define the subtasks used to decompose a task and the preconditions under which the decomposition is valid. For instance, the \textit{truckTransportM} method (in the middle of Table 1) has the same preconditions as the \textit{truckTransport} task plus the precondition that truck \textit{t} is also in location \textit{s}. In that case it decomposes the task into three subtasks. The \textit{truckTransportM} decomposition method is used twice in Figure \ref{fig:exampleFinal}.

\begin{table}[htb]
\label{table:tasks}
\caption{Some tasks and methods in the logistics transportation task domain (method names have the "M" suffix)}
\begin{tabular}{|l|l|l|l|l|}
\hline
 \textbf{Tasks/ Methods} &  \textbf{Preconditions} & \textbf{Effects/Subtasks}  \\ \hline \hline
transportPackage(p, s, d) &  at(p,s) &   at(p,d)\\ \hline
 planeTransport(p, s, d) &  at(p,s), airport(s), airport(d)&  at(p,d) \\ \hline
 truckTransport(p, s, d) &  at(p,s), sameCity(s,d)&  at(p,d) \\ \hline \hline
  truckTransportM(p, s, d) &  at(p,s), sameCity(s,d), at(t,s) &  loadTruck(t,p,s),drive(t,s,d),unloadTruck(t,p,d) \\ \hline \hline
 drive(t,s,d) & at(t,s), sameCity(s,d) & at(t,d)\\ \hline
loadTruck(t,p,l) & at(t,l), at(p,l) & at(p,t)\\ \hline
\end{tabular}

\end{table}

The other kinds of tasks are called \textit{primitive}. They have associated actions for transforming the world state. The last two rows of Table 1 list two primitive tasks and the preconditions and effects of their associated actions. For instance, the action associated with \textit{loadTruck(t,p,l)} requires truck \textit{t} and package \textit{p} to be at location \textit{l}. After the action is executed,  \textit{p} will be inside \textit{t}.

Formally, we use $T$ to denote the collection of all compound and primitive tasks. Compound tasks have associated \textit{(preconditions, effects)} pairs whereas primitive tasks have associated actions, which have their own \textit{ (preconditions, effects)} pairs.

An action $a$ defines a transition function $a\!: S \rightarrow S \cup \emptyset$, where $S$ is the set of all states. If $a(s) = \emptyset$, then $a$ is not applicable in $s \in S$. Otherwise, $a(s) \in S$ is the resulting state after taking action $a$ in $s$. For instance when applying the \textit{loadTruck(truck,pck,src)} action in a state \textit{s} where \textit{at(pck,src)} holds, this will yield a state \textit{s'} that is identical to \textit{s} except that \textit{at(pck,src)} no longer holds in $s'$ and instead \textit{at(pck,truck)} holds.

A method $m$ is used to decompose a compound task $t$. A method $m\!: S \times T \rightarrow \tilde{T} \cup \emptyset$ describes its task decomposition. Here, $\tilde{T}$ is the set of all possible sequences of tasks in $T$.  If $m(s,t) = \emptyset$, then $m$ is not applicable in $(s,t)$ (i.e., cannot be used to decompose task \textit{t} in state \textit{s}). Otherwise, $m(s,t) \in \tilde{T}$ yields a sequence of tasks (i.e.,  $t$ is decomposed into $m(s,t)$). For instance, in the example in Figure \ref{fig:exampleFinal}, if applying \textit{truckTransportM(pck,ap2,dest)} in a state \textit{s} where its preconditions hold, it will yield the task list:   \textit{load(truck2,pck,ap2), drive(truck2,ap2,dest), unloadTruck(truck2,pck,dest)}.




    

   



     

\section{ChatHTN}

An HTN planning problem is defined by the 4-tuple $(s,\tilde{t},M,A)$, where $s \in S$,  $\tilde{t} \in \tilde{T}$, \textit{M} is a collection of methods, and \textit{A} is a collection of actions. Subsequently we focus on $(s,\tilde{t})$ as $M$ and $A$ remain fixed.

Figure \ref{alg:ChatHTN} displays the ChatHTN algorithm. The lines not underlined are standard for HTN planners like SHOP \cite{nau1999shop}. The underlined lines (and segments) are new additions specific to ChatHTN.

\begin{algorithm}[htb]
\caption{The ChatHTN algorithm}
\label{alg:ChatHTN}
\begin{algorithmic}[1]
    \Procedure{ChatHTN}{$s,\tilde{t}$}
\State \textbf{return} chatSeekPlan($s,\tilde{t},())$ \Comment{() is the empty plan; a plan with no actions}
\EndProcedure
\Procedure{chatSeekPlan}{$s,\tilde{t},\pi$} \Comment{$\pi$ is the plan generated so far}
\State \textbf{if} $\tilde{t} =  ()$ \textbf{then return } 
() \Comment{returns the empty plan}

\State let $\tilde{t} = (t_0, t_1,...,t_n)$
\State \textbf{if} $t_0$ is primitive \textbf{then}
\State \ \ \ \  let $a_0$ be the action associated with $t_0$
\State \ \ \ \  \textbf{if} $a_0(s) \neq \emptyset$  \textbf{then}
\State \ \ \ \ \ \ \ \ \textbf{return}  
 \textsc{chatSeekPlan}$(a_0(s),(t_1,...,t_n),\pi \cdot (a_0))$\Comment{continue with remaining tasks}
 \State \ \ \ \  \textbf{else return $\emptyset$} \Comment{return no plans}
\State \textbf{if} $t_0$ is compound \textbf{then}
\State \ \ \ \  \textbf{for} $m_0 \in M$ \textbf{do} \Comment{$M$ is the list of all methods}
\State \ \ \ \  \ \ \ \ \textbf{if} $m_0(s,t_0) \neq \emptyset$ \textbf{then}
\State \ \ \ \ \ \ \ \ \ \ \ \ \ \ let $\pi'$ =
 \textsc{chatSeekPlan}$(s,m_0(s,t_0) \cdot$ \underline{$(t^{ver}_0)$} $\cdot (t_1,...,t_n),\pi)$ \Comment{$t^{ver}_0$ is a verifier task}
\State \ \ \ \ \ \ \ \ \ \ \ \ \ \ \textbf{if} $\pi' \neq \emptyset$ \textbf{then}
\State \ \ \ \ \ \ \ \ \ \ \ \ \ \ \ \ \ \textbf{return} $\pi'$
\State \ \ \ \  \underline{let $\tilde{t_0}$ = \textsc{chatGPTQuery}$(t_0,s,A)$} 
\State \ \ \ \   \underline{let $\pi'$ =
 \textsc{chatSeekPlan}$(s,\tilde{t_0}\cdot (t^{ver}_0) \cdot (t_1,...,t_n),\pi)$} 
\State \ \ \ \  \underline{\textbf{if} $\pi' \neq \emptyset$ \textbf{then}}
\State  \ \ \ \ \ \ \ \  \underline{\textbf{return} $\pi'$}
\State \textbf{return} $\emptyset$
\EndProcedure
\Procedure{chatGPTQuery}{$t_0,s,A$}
\State let response = \textsc{chatGPT}(\lq\lq...generate a decomposition ... " + $t_0$ + $s$ + $A$)
\State let tasks = \textsc{chatGPT}(\lq\lq... generate a sequence of primitive tasks..." + response + $t_0$ + $s$ + $A$)
\State \textbf{return} tasks
\EndProcedure
\end{algorithmic}
\end{algorithm}

ChatHTN calls the chatSeekPlan procedure with the state $s$, the task list $\tilde{t}$, and a third parameter which is the plan (a list of actions) that will be generated, initially empty (Line 2). If the task list is empty, chatSeekPlan returns the empty plan (Line 5).\footnote{If chatSeekPlan returns $\emptyset$, it means no solution was found. If it returns $()$, then a solution was found: the empty plan.} The rest of the algorithm is divided into three cases:

\begin{enumerate}
    \item (Case 1) if $t_0$ is primitive and its associated action $a_0$ is applicable in $s$ (i.e., $a_0(s) \neq \emptyset$), then continue with the new state $a_0(s)$, the rest of the tasks $(t_1,...,t_n)$, and the plan generated so far $\pi \cdot (a_0)$ (Lines 7-10). The $\cdot$ operator is list concatenation. Line 11 returns the empty set (i.e., no plans) if $a_0$ is not applicable in $s$.
     \item (Case 2) if $t_0$ is compound, it searches for a method $m_0$ applicable to $s$ and $t_0$ (i.e., $m_0(s,t_0) \neq \emptyset$) and continues with the same state $s$, the updated task list $m_0(s,t_0) \cdot$ \underline{$(t^{ver}_0)$} $\cdot (t_1,...,t_n)$, and the same plan $\pi$ (Lines 12-15). If a plan $\pi'$ is found in the recursive call, it is returned (Lines 16-17). Here is a first change with respect to standard HTN planning: the verifier task $t^{ver}_0$ is concatenated to the task list, after  $m_0(s,t_0)$. We will elaborate on verifier tasks below.
     \item (Case 3) Standard HTN continues with Line 22, returning no plans since if it reaches that line, then Cases 1 and 2 have both failed. ChatHTN instead calls chatGPTQuery, which returns a sequence of primitive tasks $\tilde{t_0}$ for $t_0$ (Line 18). ChatHTN continues with a recursive call with the updated task list (Line 19). If it succeeds, then plan $\pi'$ is returned (Line 21). If not, then it returns no plan (Line 22).
     
\end{enumerate}

The procedure ChatGPTQuery is shown in Lines 24-28. It first prompts ChatGPT to generate a task decomposition for $t_0$ and for context it also provides $s$ and $A$. Then it performs prompt chaining \cite{genkina2024ai} by querying ChatGPT to match the response it provided in Line 25 to a sequence of primitive tasks. In the second query we include the parameters given in the original query for context. Recurrent use of ChatGPT (and other chatbots) have shown that providing piecemeal queries provides better results than a single query with all requirements \cite{kwak2024classify}. We also observed this when testing our system. The exact prompts we use in ChatHTN can be found in Appendix A.

\section{Properties and Other Considerations}

ChatHTN is sound. Informally, the tasks in $\tilde{t}$ are satisfied by the plan generated by ChatHTN$(s,\tilde{t})$. 
Formally, let $\Pi$ be an HTN planner. Let $\pi = \Pi(s,\tilde{t}) = (a_0, a_1,..,a_m)$. 
We say that a task list $\tilde{t}$ of \textbf{compound} tasks is satisfied by $\Pi(s,\tilde{t})$ if the following recursive conditions are true:\footnote{This definition can be extended when $\tilde{t}$ consists of both compound and primitive tasks. We focus on compound tasks because it is both the interesting case and to simplify the definition.}

\begin{itemize}
    \item (base case) If $\tilde{t} = ()$ and $\Pi(s,\tilde{t}) = ()$, then $\tilde{t}$ is satisfied by $\Pi(s,\tilde{t})$.
    \item (base case) If $\tilde{t} = (t_0)$, and $s' = a_m(...(a_1(a_o(s)))...) \neq \emptyset$, then $(t_0)$ is satisfied by $\Pi(s,(t_0))$ if the effects of $t_0$ are satisfied in $s'$.
    \item (recursive case) If $\tilde{t} = (t_0, t_1,...,t_n)$, then let $\pi_0 = (a_0...a_x)$ (with $x \leq n$) be the prefix plan of $\pi$ generated by $\Pi(s,(t_0))$ after recursively decomposing $t_0$ into primitive tasks. Then $\tilde{t}$ is satisfied by $\Pi(s,\tilde{t})$ if:
    \begin{itemize}
    \item $(t_0)$ is satisfied by $\Pi(s,(t_0))$, and
    \item  $(t_1,...,t_n)$ is satisfied by $\Pi(s',(t_1,...,t_n))$, where $s' = a_X(...(a_1(a_o(s)))...) \neq \emptyset$.
    \end{itemize}
\end{itemize}

Under these conditions, $\tilde{t}$ is satisfied by $ChatHTN(s,\tilde{t})$. The reason for this is the use of verifier tasks in ChatHTN. A verifier task $t_0^{ver}$ for a compound task $t_0$ is a primitive task whose associated action has no effects and has as preconditions the effects of $t_0$. Therefore, when applying a verifier task  $t_0^{ver}$, the state does not change but it is inapplicable if the effects of $t_0$ are not valid in the state. Like any other primitive task, they are checked in Line 9 and if the preconditions are not valid, chatSeekPlan returns the empty set (line 11). So planning with the rest of the tasks will continue (Line 10) only if the verifier tasks are satisfied. 

We added verifier tasks in the two situations when a task is decomposed in ChatHTN. First, in Line 15, when adding the subtask decomposition $m_0(s,t_0)$ for task $t_0$, this is followed by adding $t_0^{ver}$ to the task list. This forces ChatHTN to check task $t_0$'s effects after all tasks in $m_0(s,t_0)$ are achieved. We do the same in Line 19 for the task decomposition $\tilde{t_0}$, which was generated by ChatGPTQuery. Standard HTN planning  does not include verifier tasks as tasks do not have semantics (e.g., preconditions and effects). We introduced task verifiers in \cite{hogg2008htn} in the context of automated learning of HTN methods. When reusing learned methods from a single plan trace, the resulting plans were always correct, satisfying the tasks' effects. However, as more methods were learned from other traces, they interacted in unexpected ways, resulting in plans that sometimes do not satisfy the given task. To our surprise, we experience an analogous phenomenon when ChatGPTQuery introduces its own task decompositions, even when most of the task decompositions were generated by our well-crafted methods. By \textit{well-crafted}, we mean that if ChatGPTQuery is not called, then the plans will be correct even without adding the verifier tasks. This is why we introduce task verifiers for all task decompositions, including those that are generated from manually-crafted methods, a departure from our previous work on learning HTN methods \cite{hogg2008htn}.

An HTN planner $\Pi$ is \textit{complete} if, whenever a solvable problem {($s,\tilde{t}$)} is given, then  $\Pi(s,\tilde{t})$ returns a plan. 
 HTN planners such as SHOP are not complete. Therefore, ChatHTN is also not complete. For instance, when solving a task, a method might be chosen that results in an infinite loop, whereas selecting a different method might yield a solution. Related to this point, we found that ChatHTN could enter infinite loops due to tasks introduced by ChatGPTQuery. To address this, our implementation of ChatHTN tracks previously visited \textit{(state, task)} pairs. If such a pair is encountered again, the algorithm returns an empty set (this mechanism is not explicitly shown in  Algorithm \ref{alg:ChatHTN}).

\section{Empirical Study}

We implemented ChatHTN using the PyHop HTN planner \cite{Nau2020} and released our implementation on GitHub\footnote{\url{https://github.com/hhhhmmmmm02/ChatHTN}}. We tested it on three  domains that are amenable to hierarchical task decomposition: (1) logistics transportation, (2) a household robot domain, and (3) a search-and-rescue domain. All domains are also released in the GitHub repository. These studies focus on demonstrating our theoretical result regarding the soundness of ChatGPT.

The household robot domain requires a robot to navigate between rooms in a house and perform two distinct tasks: cleaning and organizing them. A key difference from  logistics is that the task verifiers for some of the compound tasks, e.g., \textit{houseIsClean(house)},  requires  ChatHTN to check if a condition defined by an axiom is satisfied, which depends on multiple state conditions. For instance,  "\textit{houseIsClean(house)} is satisfied if  every room in the house is clean". We also use axioms to define preconditions, such as the \textit{sameCity} condition in Table 1.

The search-and-rescue domain involves a drone scanning  areas to check for potential survivors after a calamity in a given region. If it finds survivors in a particular area, it needs to relocate them to a safe location. This domain also requires checking whether conditions defined by axioms are met.

For each domain we conducted the same test. We generated a prototypical problem, one that requires decomposing all possible tasks in the domain. For the logistics transportation domain, this yields a hierarchy that is similar to the one illustrated in Figure \ref{fig:exampleFinal}, with small variations such as a truck relocating before loading a package. We performed the following tests:

\begin{itemize}
    \item {\bf Full Domain}: We ran ChatHTN on the problem with all tasks, methods, and actions to ensure it is solvable.
    \item \textbf{Unsolvable}: We removed a condition from the problem, making it unsolvable. We then ran ChatHTN on the problem with all tasks, methods, and actions as a demonstration of the theoretical property that it will \textit{not} generate a solution because ChatHTN generates only correct plans.
    \item {\bf Method Removed}: For each compound task in the domain, we removed one method for decomposing the task while leaving the others available for ChatHTN. We tested whether it could still generate a plan. We repeated this test for every method of every task.
    \item {\bf No Methods}: We removed all methods for a single compound task and attempt to solve the problem with ChatHTN, which has access to the other tasks and their methods. We repeated this test for each task in the domain.
    \item {\bf No Model}: We removed all methods from all tasks to test whether ChatHTN could still generate a plan. This is akin to planning with LLMs, with the only difference being that we test whether the top-level task's effects are still valid (e.g., \textit{transportPackage} for the transportation domain).
\end{itemize}

One observation from our tests is that sometimes ChatHTN will fail to generate a solution for a given solvable problem whereas other times it generates a solution for that same problem. This occurs because ChatGPT rarely generates the same output when prompted with the same input.  Therefore, for each of our tests, we gave ChatHTN five chances to solve a given problem. This happened even if we set ChatGPT's parameter \textit{temperature} to zero, which reduces randomness. In our experiments, we use the default temperature, which is one. This enables ChatGPT to explore alternatives while   still using a coherent model.

\begin{table}[htb]
\centering
\caption{Results for the three domains. An \lq\lq X" indicates that no solution was generated and a check mark denotes that a solution was generated.  The compound tasks are listed in italics. The number in parentheses indicates the number of calls to \textsc{chatGPTQuery}. Thus, 0 means no calls, whereas 4 indicates four calls. If there is more than one number in parentheses next to a checkmark, then it indicates the number of runs until a solution was generated. For instance, \textit{(5,2)}, indicates (1) the first time it failed and there were 5 calls and (2) the second time it succeeded and there were only 2 calls.}
\begin{tabular}{lllllll}
 \textbf{Logistics} &  & \textbf{Household robot} &  & 
 \textbf{Search and rescue} & \\
 Full domain& \checkmark(0) & Full domain & \checkmark(0) &  Full domain & \checkmark(0) &\\
 Unsolvable &  X(4) & Unsolvable  &  X(3)&  Unsolvable  &  X(4)&\\
\textit{truckTrans.}& \checkmark(5,2) & \textit{sweepTask} &  \checkmark(1) & \textit{scanAreaTask} & \checkmark(1) &\\
M1& \checkmark(0)  & M1 & \checkmark(0) & M1 & \checkmark(0) &\\
M2& \checkmark(2)  & M2 & \checkmark(1) & M2 & \checkmark(0) &\\
M3& \checkmark(5,2)  & \textit{org.Task} & \checkmark(1) & M3& \checkmark(1)  &\\
\textit{planeTrans.}& \checkmark(1) & M1 & \checkmark(1) & \textit{rescueSurvivor}& \checkmark(2)&\\
M1& \checkmark(0)  & M2 & \checkmark(1) & M1 & \checkmark(2) &\\
M2& \checkmark(1)  & \textit{cleanHouse} & \checkmark(1) & M2 & \checkmark(1) &\\
M3& \checkmark(1)  & M1 & \checkmark(1) & \textit{rescueSurvivor} & \checkmark(2) &\\
\textit{trans.Pack.}&  \checkmark(1,1,1,1) & M2  & \checkmark(1) & \textit{checkSurvivors}& \checkmark(1) & \\
M1& \checkmark(0)  & \textit{org.House} & \checkmark(2) & M1& \checkmark(0)  &\\
M2& \checkmark(1,1,1,1,1)  &  M1&  \checkmark(1) & M2& \checkmark(1)  &\\
&  & M2 & \checkmark(2) & \textit{search and rescue}&  \checkmark(1,1)&\\
&   & \textit{careHouse} & \checkmark(1) & M1&  \checkmark(1)&\\
&   &  M1& \checkmark(1) & M2& \checkmark(0)&\\
&   &  &  & M3& \checkmark(1)&\\
\end{tabular}

\label{results}
\end{table}

The results of our tests are shown in Table \ref{results}. We confirmed that, when the {\bf Full Domain} was provided, the task was solvable and no calls were made to ChatGPT. Also as expected, ChatHTN failed to find a solution for the {\bf Unsolvable} test. For each domain, we then list the compound tasks, starting with the simplest first. The final task listed per domain is the top level task, one that when using \textbf{Full Domain} will result in a hierarchy that includes all other compound tasks.  

Sometimes a solution could be generated without calling chatGPTQuery despite a particular method being hidden. The reason for this is  that those methods cover  corner cases such as when the task is already solved (e.g., the effects are satisfied in the state). Another observation from our tests is that when a problem is unsolvable or the top-level task fails then chatGPTQuery will still generate a sequence of tasks, even though this sequence is not a correct solution. In each such case, ChatHTN will catch the error because of its verifier tasks and for which we annotate an "X" in its corresponding table entry.

At the time of this writing and the current pricing for using ChatGPT's version gpt-4-turbo for each domain, it took around \$30 per domain for tuning and running all the tests. This version of ChatGPT is considered cost-effective. Other versions are more powerful but are also much more costly. If we used a more powerful version of ChatGPT (or other chatbot), presumably more complex domains could be tested, with more complex prompt chaining designs (e.g., the hierarchy generated so far could be given as input to ChatGPT). However, we leave that as a topic for future research.

\section{Related Work}

There has been substantial work on learning HTNs including learning methods' preconditions \cite{zhuo2009learning}, learning methods (i.e., both the preconditions and effects) \cite{langley2009learning, hogg2008htn}, and learning methods and operators \cite{zhuo2014learning}. Among these prior efforts, the most closely related is the one using teleoreactive learning \cite{li2011improving}. In their work, when the algorithm is performing HTN planning and the system finds a gap in its knowledge base, it reverts to first-principles planning to generate a plan to fill the gap. It then learns methods from the generated plan. In future work, we could do something similar by learning methods from the decompositions generated by ChatGPTQuery. 

Efforts on combining planning and LLM have also gained substantial attention. Studies have shown that, while LLMs cannot generate nor verify plans, they could be used to generate approximate solutions \cite{valmeekam2024planningstrawberryfieldsevaluating}. This is a motivation for our work; to provide soundness guarantees despite using ChatGPT to approximate parts of a solution.

Work has also been conducted on generating a hierarchical planning structure from instructions given to LLMs. This has been done in combination with temporal linear logic to execute the generated hierarchies \cite{luo2023obtaining}. In our work, we go in the opposite direction; we generate hierarchies using HTN planning and generate localized task decompositions using LLMs while reverting back to HTN planning when possible. This is done for efficiency reasons and for reliability as the knowledge base has been carefully curated.

Work has explored possible uses of LLMs in combination with hierarchical planning \cite{puerta2025roadmap}. This indicates that LLMs could be used to elicit the problem's description from the user, help generate the plan, and perform post-processing once a plan is generated. It is possible in future work that we could use LLMs to elicit the tasks that the user wants to achieve (e.g., to select among the compound tasks defined in the knowledge base).

An example of the latter point is explored in \cite{ding2023task}. The user gives as input a plain language description of the problem. This work is in in the context of underwater unmanned vehicles that need to explore ocean regions. It uses similarity functions to identify which of the tasks the user is aiming to achieve and then proceeds to perform hierarchical planning.

\section{Final Remarks}

In this paper we introduced ChatHTN, a sound planning algorithm to solve HTN planning problems. At its core ChatHTN is a symbolic HTN planner that performs task decomposition. However, it prompts ChatGPT with queries whenever it encounters situations where there is no applicable method to decompose a task. It uses ChatGPT's approximate planning capabilities, when needed, to generate a plausible decomposition. It then reverts control back to symbolic HTN planning. Our work  leverages sound HTN planning typical of symbolic planners while alleviating the stringent requirement that a complete knowledge base must be provided apriori that covers all  cases. It ensures that the plans which are generated are correct while relaxing the requirement that a complete collection of methods must be provided. 

There are several potential lines of future research. We would like to expand and test ChatGPT on more complex tasks, including some that may require a hierarchy with dozens of levels. We would also like to test ChatGPT in domains that are not from the planning literature yet are still amenable to hierarchical decomposition. One possible way to tackle these more complex scenarios is to relax the requirement that  \textsc{chatGPTQuery} returns a sequence of primitive tasks, Instead,  we plan to enable \textsc{chatGPTQuery} to generate a sequence of primitive and compound tasks. Finally, in some of the problems \textsc{chatGPTQuery} is queried repeatedly on the same decomposition. There is an opportunity to learn these decompositions as HTN methods, thereby reducing the calls to \textsc{chatGPTQuery} over time.


\bibliography{sample-old}

\section*{Appendix A: ChatGPT prompts}

In this appendix, we show the exact prompts used by ChatGPT. These prompts are parametrized so that when the domain changes, we pass the domain-specific information such as actions, tasks, etc., while the rest of the prompt remains the same.

The following is the prompt used for ChatGPT in Line 25 of the Algorithm \ref{alg:ChatHTN}. The "+" operator below is text concatenation. Here we are passing text versions (automatically generated by processing the PyHop's knowledge base) of the task, the task's preconditions (precs), the task's effects (effs), state, actions (operatorsText), and axioms. The latter are auxiliary functions to simplify the  preconditions. For instance, \textit{samecity(loc,loc2)} is a an auxiliary function used by the \textit{drive} action to check that the starting and ending locations are in the same city.

\textit{
conversation=[\\
        \{"role": "system", "content": "You are an AI planner specializing in HTN planning."\},\\
        \{ "role":"user", 
        "content": "The domain is defined by the following operators (each defined as a  Python function):" + operatorsText +
        ". Some of the preconditions in the operators are defined by the following python functions: " +axiomsText+
        ". Provide the Sub-Tasks Breakdown for the following task: " + task  +
        ". Here are the preconditions of the task: " + precs +
          ". Here are the effects of the task: " + effs +
        ". Here is the current state: " +stateText + 
        " Provide a complete and logically valid decomposition using the operators and functions provided."+
        " Do not invent new operators. Your output should be a step-by-step list of sub-tasks in logical order,"+
        " using arguments ground in the current state."\}
    ]
}

    The following is the second prompt after the response to the previous prompt has been generated (Line 26 of Algorithm \ref{alg:ChatHTN}). The variable \textit{response} is the response by ChatGPT to the previous prompt. 

\textit{
conversation=[\\
     \{"role": "system", "content": "You are an AI planner specializing in HTN planning."\},\\
        \{ "role":"user", 
        "content": 
        ". You generated the following response:"+chatGPTtext+ 
        "to my request to provide the Sub-Tasks Breakdown for the following task: " + task + 
        ". I also gave you the preconditions of the task: " + precs + " and the effects of the task: " + effs+
        ".and gave you the state: " +state + 
        "and gave you the domain  defined by the following operators (each defined as a  Python function):" + operatorsText+
        ". I also gave you the following python functions which are called to check some preconditions and to check some effects: " +axiomsText+
        ". As a follow-up, can you map  the subtasks you generated with the operators I provided, please? please list the operator names as predicates,"+
        " for the match you generate use only the predicate names of the operators and the arguments in your sub-task breakdown"+
        " Always respond with a compact, machine-readable format using predicate form. "+
        " Avoid explanations or extra text unless explicitly requested. "+
        " When generating your output, list only the predicates in the form: "+
        " predicate(arg1, arg2, ...)"+
        " where predicate is name of an operator I provided"+
        " Separate predicates by newlines. Do not include explanations, headings, or descriptions." +
        " Use only the operator names I provided. Ensure that every predicate corresponds exactly"+
        " to one of those operators and that all arguments match those in your sub-task breakdown"\}
    ]
}

\end{document}